\documentclass[letterpaper]{article} 
\usepackage{aaai25}  
\usepackage{times}  
\usepackage{helvet}  
\usepackage{courier}  
\usepackage[hyphens]{url}  
\usepackage{graphicx} 
\usepackage{natbib}  
\usepackage{caption} 
\usepackage{algorithm}
\usepackage{algorithmic}
\usepackage{kotex}
\usepackage{lipsum}
\usepackage{multirow}
\usepackage{hhline}
\usepackage{subcaption}
\usepackage{newfloat}
\usepackage{listings}
\DeclareCaptionStyle{ruled}{labelfont=normalfont,labelsep=colon,strut=off} 
\floatstyle{ruled}
\newfloat{listing}{tb}{lst}{}
\floatname{listing}{Listing}
\usepackage{amsmath} 
\usepackage{diagbox} 
\usepackage{colortbl} 
\usepackage{xspace} 
\usepackage{kotex} 
\usepackage{booktabs} 
\usepackage{hyperref} 
\usepackage{footmisc} 
\nocopyright 

\newcommand{\fref}[1]{Figure~\ref{#1}}
\newcommand{\eref}[1]{Eq.~(\ref{#1})}
\newcommand{\tref}[1]{Table~\ref{#1}}
\newcommand{\sref}[1]{Section~\ref{#1}}
\newcommand{\aref}[1]{Appendix \ref{#1}}
\newcommand{\loss}{point-wise semantic loss\xspace}
\newcommand{\losscap}{Point-wise semantic loss\xspace}
\newcommand{\ours}{Ours}
\newcommand{\prevseg}{rendered 2D segmentation\xspace}

\newcommand{\PREVSEG}{Rendered 2D Segmentation\xspace}
\usepackage{pifont}
\newcommand{\cmark}{\ding{51}} 
\newcommand{\xmark}{\ding{55}} 
\newcommand{\pos}{\mathbf{x}}  

\title{Rethinking Open-Vocabulary Segmentation of Radiance Fields in 3D Space}
\author {
    Hyunjee Lee\equalcontrib,
    Youngsik Yun\equalcontrib,
    Jeongmin Bae,
    Seoha Kim,
    Youngjung Uh\thanks{Corresponding author.}
}
\affiliations {
    Yonsei University\\
    \{hyunji12, bbangsik, jaymin.bae, hailey07, yj.uh\}@yonsei.ac.kr
}

\usepackage{bibentry}

\begin{document}
\maketitle


\begin{abstract}
Understanding the 3D semantics of a scene is a fundamental problem for various scenarios such as embodied agents.
While NeRFs and 3DGS excel at novel-view synthesis, 
previous methods for understanding their semantics have been limited to incomplete 3D understanding: their segmentation results are rendered as 2D masks that do not represent the entire 3D space. 
To address this limitation, we redefine the problem to segment the 3D volume and propose the following methods for better 3D understanding. We directly supervise the 3D points to train the language embedding field, unlike previous methods that anchor supervision at 2D pixels. 
We transfer the learned language field to 3DGS, achieving the first real-time rendering speed without sacrificing training time or accuracy. Lastly, we introduce a 3D querying and evaluation protocol for assessing the reconstructed geometry and semantics together. Code, checkpoints, and annotations are available at the project page.
\end{abstract}

\begin{links}
    \link{Project page}{https://hyunji12.github.io/Open3DRF/}
\end{links}


\begin{figure}[htb!]
  \centering
  \includegraphics[width=0.77\linewidth]{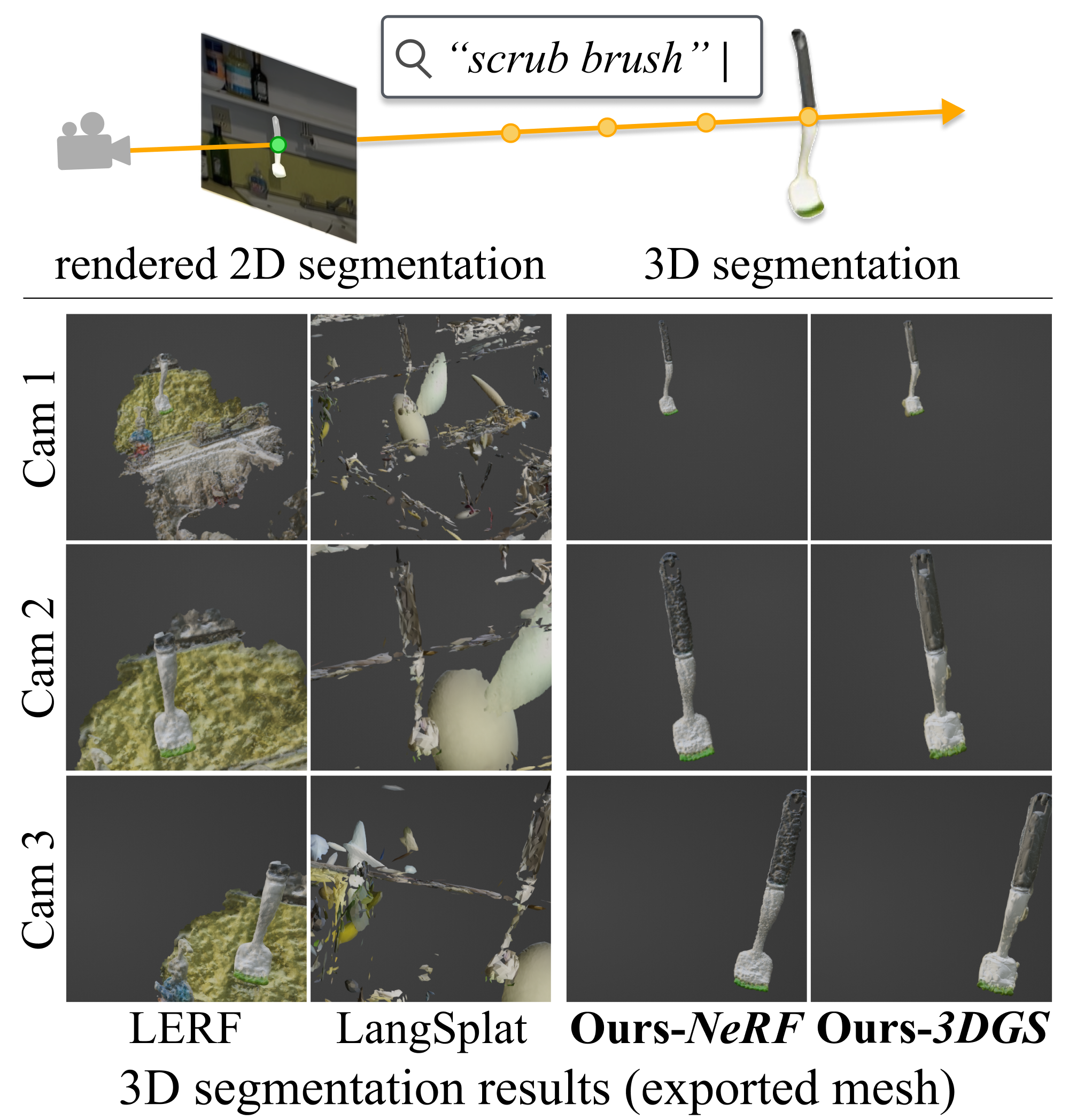}
  \caption{
  Previous works segment rendered 2D masks on rendered features to understand radiance fields. Instead, we reformulate the task to segment 3D volumes. Our approach significantly improves 3D understanding of radiance fields. 
  }
  \label{fig:teaser}
\end{figure}


\begin{figure*}[htb!]
  \centering
  \includegraphics[width=0.78\linewidth]{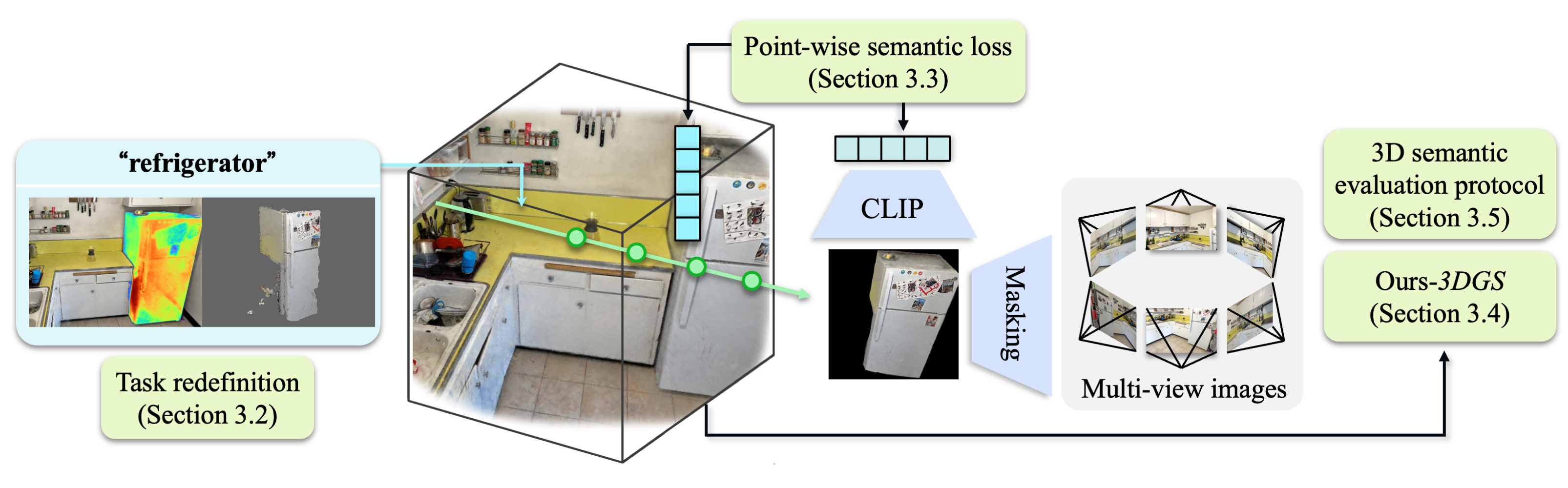}
  \caption{
  We propose 3D segmentation as a more practical problem setting, segmenting the 3D volume for a given text query (\sref{sec:querying}).
  Then we propose \loss to supervise the sampled point embeddings (\sref{sec:pploss})
  Furthermore, the learned language fields can be transferred into 3DGS for faster rendering speeds (\sref{sec:3dgs_lang}).
  Lastly, our 3D evaluation protocol measures the 3D segmentation performance both in reconstructed geometry and semantics (\sref{sec:3dseg}).
  }
  \label{fig:overview}
\end{figure*}


\section{Introduction}
\label{sec:intro}
Semantically understanding 3D space is important for various computer vision tasks. For instance, it is crucial to segment 3D objects accurately for robot manipulation~\cite{lerftogo2023,zheng2024gaussiangrasper3dlanguagegaussian}.
Recently, several works have focused on understanding 3D scenes represented by radiance fields such as Neural Radiance Fields (NeRFs)~\cite{mildenhall2020nerf} and 3D Gaussian Splatting (3DGS)~\cite{kerbl3Dgaussians}.
LERF~\cite{lerf2023} introduces a language embedding field which is rendered on a chosen viewpoint to be queried by open vocabulary.
The language field is supervised by CLIP~\cite{radford2021learning} embeddings from the multi-scale patches to capture various sizes of objects.
Subsequent works~\cite{zhang2024open,qin2023langsplat} incorporate SAM masks~\cite{kirillov2023segany} to supervise the language field for clear segmentation boundaries.

We revisit the 3D understanding of NeRFs and 3DGS in four aspects: problem setting, supervision, embeddings, and evaluation.
The problem setting of previous works leads to limited 3D understanding as they merely produce rendered 2D masks for given viewpoints rather than 3D semantic volumes. On the other hand, we set the problem to segment 3D volume regarding the semantics of 3D points. 

The previous methods encourage the rendered embeddings, rather than the embeddings in 3D space, to match the ground truth.
Furthermore, their multi-scale language embeddings require finding the optimal scale for a given viewpoint and a query text, which is prone to multi-view inconsistency. In contrast, we encourage 3D points to learn language embeddings following our revisited problem setting. Although the ground truth is still language embeddings of 2D images, changing the dimension for computing the loss greatly improves 3D understanding as shown in \fref{fig:teaser}. In addition, we remove multi-scale embeddings by encoding masked objects, achieving multi-view consistent rendered feature maps. 

Meanwhile, previous works on understanding 3DGS explicitly add language embeddings for each Gaussian and jointly train them~\cite{qin2023langsplat,zhou2023feature}. 
However, directly appending a 512-dimensional language embedding to each Gaussian and rendering them result in out-of-memory (OOM) issues. Hence, they either 1) compress language embeddings to low-dimensional features, significantly sacrificing the accuracy~\cite{shi2023language,qin2023langsplat}, or 2) modify the rasterizer, which is slow for training~\cite{zhou2023feature,ye2023mathematical}. 
To overcome these drawbacks, we transfer our learned language field into 3DGS to enable the real-time rendering of the accurate language field.

Our final aspect is evaluation. Existing works 
measure mIoU \textit{in pixels} between the rendered 2D masks and the 2D ground truth,
or measure mIoU \textit{on ground truth point clouds} ignoring the reconstructed geometry \cite{engelmann2024opennerf}.
However, 3D understanding should know the correct 3D volume of a target semantics. 
Therefore, the evaluation method should assess semantics and the reconstructed geometry together.
Accordingly, we propose to evaluate the accuracy of 3D understanding as the agreement between estimated volume in mesh and ground truth mesh, measured in F1-score. 
The proposed evaluation is applicable to both NeRF and 3DGS.

In the experiments, we demonstrate the superiority of our method regarding 1) 3D and \prevseg accuracy, 2) training and rendering time, and 3) consistency across viewpoints. 

In summary, our contributions are:
\begin{itemize}
    \item We propose a practical problem setting for 3D understanding of NeRFs and 3DGS.
    \item We propose to directly supervise 3D points before volume rendering to learn a language embedding field. It achieves the state-of-the-art accuracy in 3D and \prevseg.
    \item We propose to transfer the language field to 3DGS. It achieves the first real-time rendering speed among open-vocabulary methods which is 28$\times$ faster than the previous fastest method.
    \item We propose a 3D evaluation protocol between estimated volume and ground truth volume represented as meshes.
\end{itemize}


\section{Related Work}
In this section, we briefly review neural 3D scene representations, especially NeRF and 3DGS, and then discuss the semantic understanding of 3D scenes.

\paragraph{Radiance Fields}  
NeRF \cite{mildenhall2020nerf} reconstructs a scene as a continuous function that maps 3D points to radiance and density values. 
Volume rendering pipeline renders the points to determine the color of each pixel on an image from an arbitrary perspective. Rather than volumetric rendering, 3DGS \cite{kerbl3Dgaussians} achieves fast rendering speeds by projecting 3D Gaussians onto the camera plane followed by depth-sorted alpha-blending. Each 3D Gaussian stores location, rotation, scale, opacity, and color. 
Recent studies use these representations to semantically understand 3D scenes for various applications such as robotics~\cite{lerftogo2023,wang2023d3fields,zheng2024gaussiangrasper3dlanguagegaussian,vanoort2024openvocabularypartbasedgrasping,li2024objectaware}.

\paragraph{Semantic Understanding of Radiance Field}
The most straightforward way to understand 3D scenes represented by neural radiance fields is by adding an auxiliary branch for semantic segmentation~\cite{zhi2021place,wang2022dmnerf,liu2023weakly,gaussian_grouping}. This approach allows synthesizing semantic masks from novel views but requires a pre-defined list of target classes before training, called \textit{closed}-set. Therefore, retraining or using additional models for untrained queries is necessary, limiting application for open-vocabulary scenarios. As vision-language model (VLM) features (e.g., CLIP~\cite{radford2021learning}) expand the semantic understanding to \textit{open}-set, it becomes a widely used approach to distill CLIP features into 3D scenes. 

LERF~\cite{lerf2023} pre-computes multi-scale patches to prepare multi-scale ground-truth CLIP features. Similar to LERF, FMGS~\cite{zuo2024fmgs} pre-computes multi-scale ground-truth CLIP features and averages them to generate a low-resolution hybrid feature map for training. Instead of using patches, LEGaussians~\cite{shi2023language} distills pixel-level CLIP features. Similarly, OpenNeRF~\cite{engelmann2024opennerf} computes pixel-level CLIP features using OpenSeg~\cite{ghiasi2022scaling}. GOI~\cite{qu2024goi} trains the codebook to compress the high-dimensional semantic features into low-dimensional vectors\footnote{At the time of this work, the code of GOI was unavailable. However, its released code showed 15 fps on LERF dataset, falling short of real-time rendering.}. As it produces mixed CLIP features to contain multiple objects into a patch, LangSplat~\cite{qin2023langsplat} utilizes SAM~\cite{kirillov2023segany} to create patches for a single object. OpenMask3D~\cite{takmaz2023openmask3d} creates a bounding box from the SAM mask, and crops the image to make per-mask features. However, OpenMask3D uses bounding box, which still leads to mixed CLIP features. Since studies using multi-scale feature maps select the most relevant scale, different scales can be chosen at different views with the same query, leading to multi-view inconsistent results~\cite{lerf2023,qin2023langsplat,shi2023language}. 
On the other hand, we use SAM to make the field free from scales.

\paragraph{Language Embedded 3DGS}
Directly embedding high-dimensional CLIP features into each Gaussian is infeasible due to the limited GPU shared memory. Therefore, recent studies for understanding open-vocabulary 3D segmentation using 3D as a backbone address this problem in two ways.  LEGaussians and LangSplat compress the CLIP features using an autoencoder~\cite{shi2023language, qin2023langsplat}. However, they need to optimize the autoencoder for each scene and endure performance degradation. Others using 3DGS without feature compression ~\cite{zhou2023feature,labe2024dgd} utilize global memory, suffering for longer training time. Meanwhile, we effectively optimize 3DGS transferred from our learned language field, without the above trade-off.


\begin{figure}[tb]
  \centering
  \includegraphics[width=0.75\linewidth]{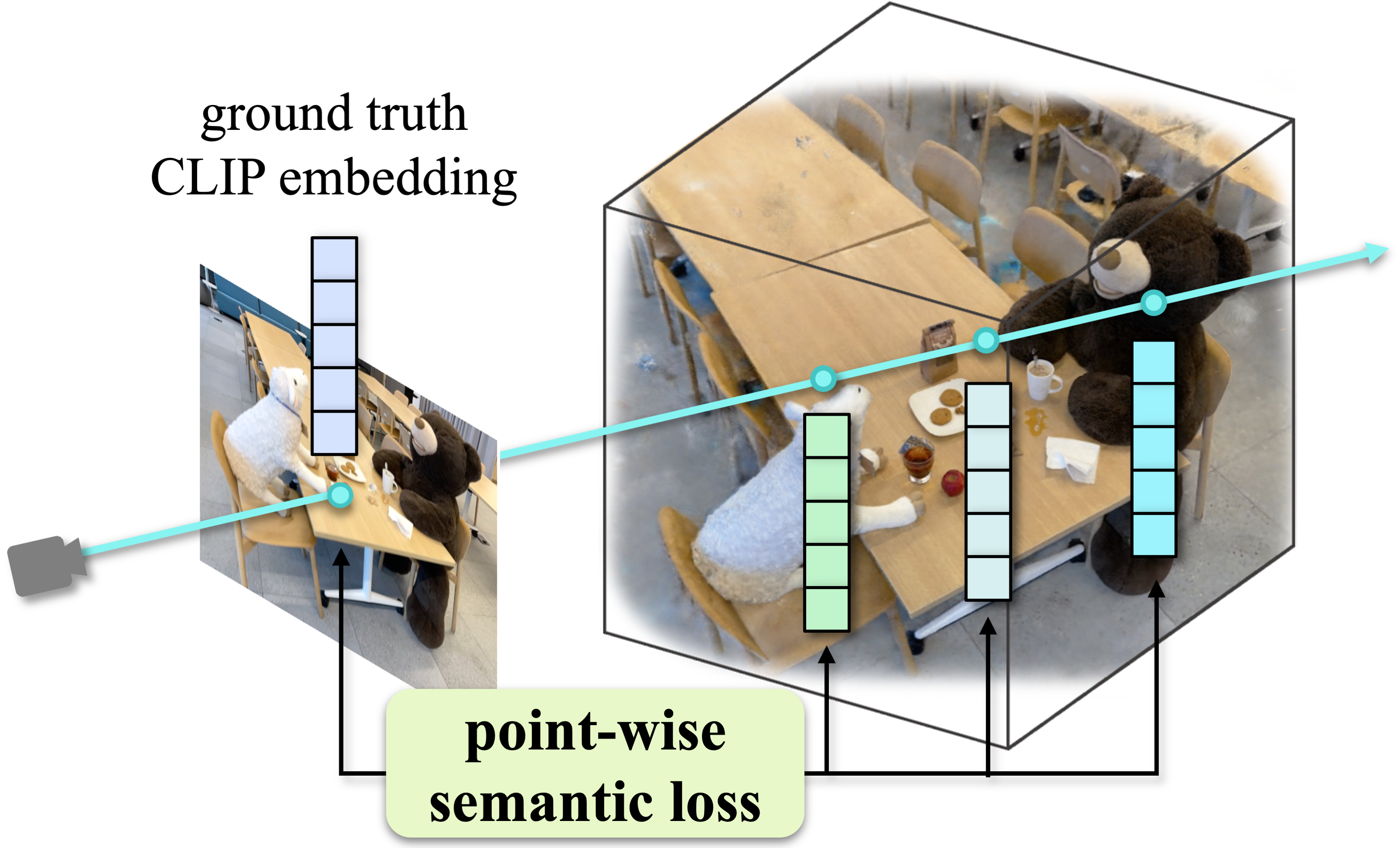}
  \caption{\losscap
  supervises the language embeddings of sampled points directly in 3D space, ensuring precise semantics.
  }
  \label{fig:pploss}
\end{figure}


\section{Methods}
First, we provide preliminary on LERF \cite{lerf2023}. Then we redefine 3D segmentation of radiance fields as computing 3D relevancy scores. Accordingly, we introduce \loss to supervise ray points in 3D space by our scale-free embeddings.
In addition, we propose to transfer our learned language field into 3DGS for real-time rendering. Lastly, we introduce the evaluation protocol for 3D segmentation. \fref{fig:overview} illustrates an overview.

\subsection{Preliminary: LERF} 
\label{sec:prelim}
For open-vocabulary segmentation, LERF builds an additional language field on top of iNGP~\cite{mueller2022instant}. The language field $F_{\text{lang}}$ is jointly trained with the radiance field by querying point embeddings $F_{\text{lang}}(\pos, s)$ at $N$ sample points' position $\pos$ and scale $s$ along each ray. 
The 3D point language embeddings are then accumulated via the volume rendering~\cite{max1995optical} to obtain rendered language embedding $\hat{\phi}^s_{\text{lang}}$ in 2D pixel space: $ \hat{\phi}^s_{\text{lang}} = \sum_{i=0}^N w_iF_{\text{lang}}(\pos_i, s)$ .
The weight of each sampled point is calculated as $ w_i=T_i(1-\exp(-\sigma_i\delta_i) \:$, where $T_i=\exp(-\sum_{j=1}^{i-1} \sigma_j\delta_j)$ is transmittance, $\delta$ is the distance between adjacent samples, and $\sigma$ is the volume density.
Then, the rendered embedding is normalized to the unit sphere as in CLIP: $\phi_{\text{lang}}^s=\hat{\phi}_{\text{lang}}^s/||\hat{\phi}_{\text{lang}}^s||$
To train the language field, LERF crops the training dataset into multi-scale patches, creating ground truth language embedding $\phi_{\text{lang}}^{\text{gt}_s}$ and maximizing the cosine similarity between the rendered language embedding $\phi_{\text{lang}}$:
\begin{equation}
    L_{\text{lang}} = - \sum_s \lambda_{\text{lang}}\: \phi_{\text{lang}}^s \cdot \phi_{\text{lang}}^{\text{gt}_s}.
    \label{eq:2dloss}
\end{equation}

Furthermore, LERF builds an additional branch for the DINO \cite{caron2021emerging} feature field as an extra regularizer for achieving clearer object boundaries.
Similar to the above, the DINO branch is trained to maximize the similarity between rendered DINO embedding $\hat{\phi}_{\text{dino}}$ and the DINO ground truth $\phi_{\text{dino}}^{\text{gt}}$. The DINO branch is not used during inference.
 
For \prevseg, LERF computes the relevancy score using the rendered embeddings $\hat{\phi}_{\text{lang}}^s$ and obtains the 2D mask:
\begin{equation}
    \min\limits_{i} {\frac{\exp(\hat{\phi}_{\text{lang}}^s \cdot \phi_{\text{text}})}{\exp(\hat{\phi}_{\text{lang}}^s \cdot \phi_{\text{text}}) + \exp(\hat{\phi}_{\text{lang}}^s \cdot \phi_{\text{canon}}^i)}}. 
    \label{eq:2dqry}
\end{equation}


\subsection{Task Redefinition}
\label{sec:querying}

Existing methods provide limited 3D semantic understanding for given texts via \textit{rendered} 2D masks on viewpoints that do not directly represent the entire 3D space. Instead, we propose \textit{3D segmentation} as a more practical problem setting: segmenting the 3D volume for given texts. 
While following the basic elements of language embeddings and relevancy scores, we compute relevancy scores of the language embeddings queried \textit{on 3D points} $\pos$ instead of the ones rendered on 2D images:
\begin{equation}
    \min\limits_{i} {\frac{\exp(F_{\text{lang}}(\pos) \cdot \phi_{\text{text}})}{\exp(F_{\text{lang}}(\pos) \cdot \phi_{\text{text}}) + \exp(F_{\text{lang}}(\pos) \cdot \phi_{\text{canon}}^i)}}, 
    \label{eq:3dqry}
\end{equation}
where $\phi_{\text{canon}}^i$ represents predefined canonical texts such as \texttt{photo} and \texttt{image}.
In NeRFs, we compute the relevancy scores on the points along the rays through pixels.
For 3DGS, we compute the relevancy scores on the center positions of the Gaussians. 
The regions where the computed relevancy scores surpass the selected threshold are considered object regions.


\subsection{Supervising Semantics in 3D Space}
\label{sec:pploss}

It is a reasonable choice to minimize the error between the rendered colors and the ground truth colors on 2D images for novel view synthesis.
In contrast, the similar objective for language embeddings (\eref{eq:2dloss}) harms correctly understanding 3D semantics as shown in \fref{fig:teaser}.

To address this issue, we propose a \loss~which uses CLIP embeddings $\phi_{\text{lang}}^{\text{gt}}$ as direct ground truth for the embeddings of 3D points on the ray $F_{\text{lang}}(\pos)$.
Specifically, we maximize the similarity between point embeddings $F_{\text{lang}}(\pos)$ and the ground truth CLIP embedding $\phi_{\text{lang}}^{\text{gt}}$:
\begin{equation}
    L_\text{PS} = -\sum_{i=0}^N (w_i F_{\text{lang}}(\pos_i) \cdot \phi_{\text{lang}}^{\text{gt}}).
\label{eq:3dloss}
\end{equation}

We also apply the same approach described above for DINO regularization. We show that \loss improves not only 3D segmentation but also \prevseg thanks to a better understanding of 3D semantics. 

Notably, multi-scale approaches~\cite{lerf2023,qin2023langsplat} lead to multi-view inconsistency because these approaches have language embedding per scale and different views may have different optimal scales.
To address this, we use SAM~\cite{kirillov2023segany} to obtain object masks and create scale-free ground truth CLIP embeddings by masking out their background and cropping tightly to the masks. It ensures view-consistent language fields without the need to determine the optimal scale. We note that previous methods with SAM still model multi-scale CLIP embedding fields \cite{zhang2024open,qin2023langsplat}.

\begin{figure}[tb]
  \centering
  \includegraphics[width=0.95\linewidth]{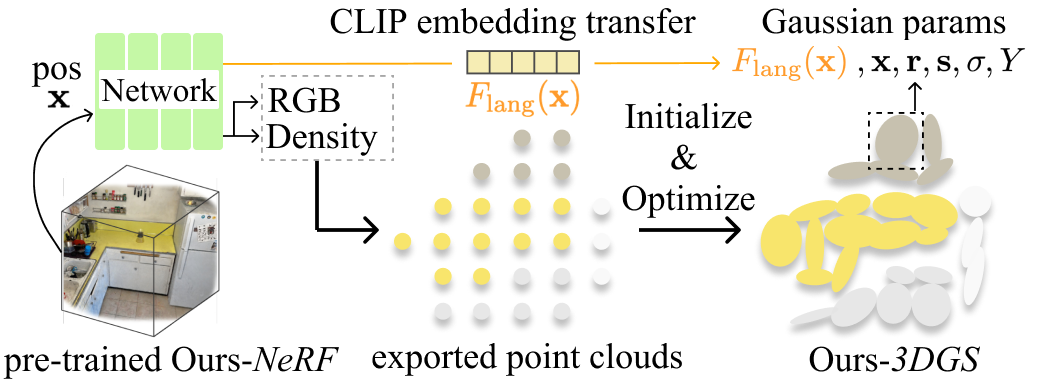}
  \caption{Transferring \ours-\textit{NeRF} into 3DGS: We initialize 3DGS using the point cloud exported from our learned NeRF, then optimize the attributes of 3DGS except for position. The language features obtained by querying the language field at the Gaussian center positions are then transfer to 3DGS. 
  }
  \label{fig:ours-3dgs}
\end{figure}

\subsection{Transferring Language Field into 3DGS}
\label{sec:3dgs_lang}

A straightforward approach for the same task with 3DGS is to jointly optimize the Gaussians and their additional language embeddings, which requires vast memory consumption. It suffers from slow training\footnote{Training on GPU global memory is slow due to the gradient computation in the backward pass. The forward pass is still fast.}~\cite{zhou2023feature}, or requires an additional model for feature compression, which degrades accuracy due to compression loss~\cite{shi2023language,qin2023langsplat}.
Appendix provides more details.

To tackle this problem, we propose to simply bake our learned language field into 3DGS by querying the language embedding at the center of the Gaussians. The queried embeddings are frozen. It runs instantly (22ms) and still allows fast rendering\footnotemark[2].

In addition, we propose to initialize the center coordinates of 3DGS from the learned radiance field.  
We collect ray points and extract the top 1M points regarding density following NeRFstudio~\cite{nerfstudio}. 
Unlike previous methods~\cite{niemeyer2024radsplat}, we implement this approach to align the geometry of 3DGS with the language field. Therefore, we freeze the center position of the Gaussians $\pos$, without densification or pruning during training. 
We then transfer language embeddings $F_{\text{lang}}(\pos)$ at the center of Gaussians. 

Similar to previous methods, we modify the rasterizer to render high-dimensional language features (see Appendix).
The rendered language embedding $\hat{\phi}_{\text{lang}}$ is obtained by $\alpha$-blending:
\begin{equation}
    \hat{\phi}_{\text{lang}} = \sum_{i \in N}F_{\text{lang}}(\pos_i) \alpha_i \prod_{j=1}^{i-1}(1-\alpha_j),
    \label{eq:rasterize}
\end{equation}
where $N$ denotes the number of Gaussians overlapping on the pixel.
The alpha value is calculated as $\alpha_i = \sigma_iG(\pos_i),$ where $G$ denotes the Gaussian kernel and $\sigma$ denotes the opacity of Gaussians. We use \eref{eq:3dqry} for 3D segmentation.


\begin{figure}[tb]
  \centering
  \includegraphics[width=0.7\linewidth]{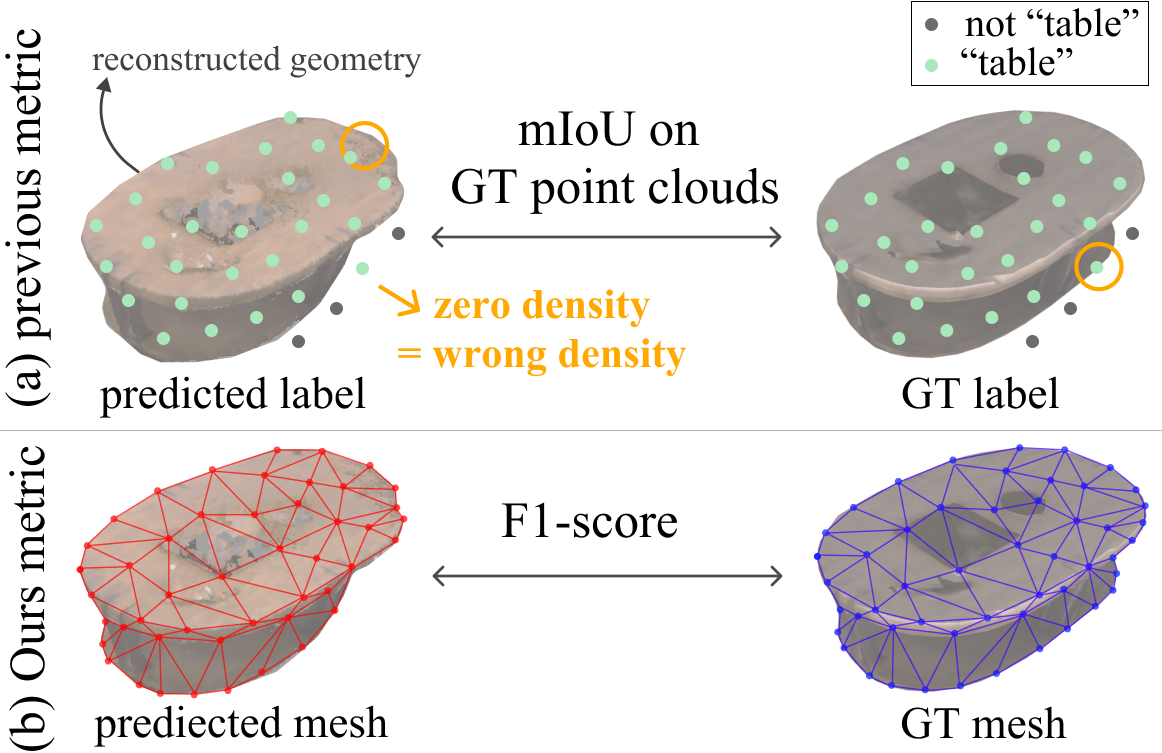}
  \caption{Comparison of 3D Evaluation:  
(a) Existing methods predict the labels at the ground truth point cloud.
It is misleading when the language embeddings capture the object area while the reconstructed geometry does not cover that object area.
(b) To address this problem, we extract 3D meshes from the segmented points of the reconstructed scene to measure the F1-score between the exported mesh and the ground truth mesh.
  }
  \label{fig:3d-seg-method}
\end{figure}

\begin{figure*}[tb]
  \centering
  \includegraphics[width=0.7\linewidth]{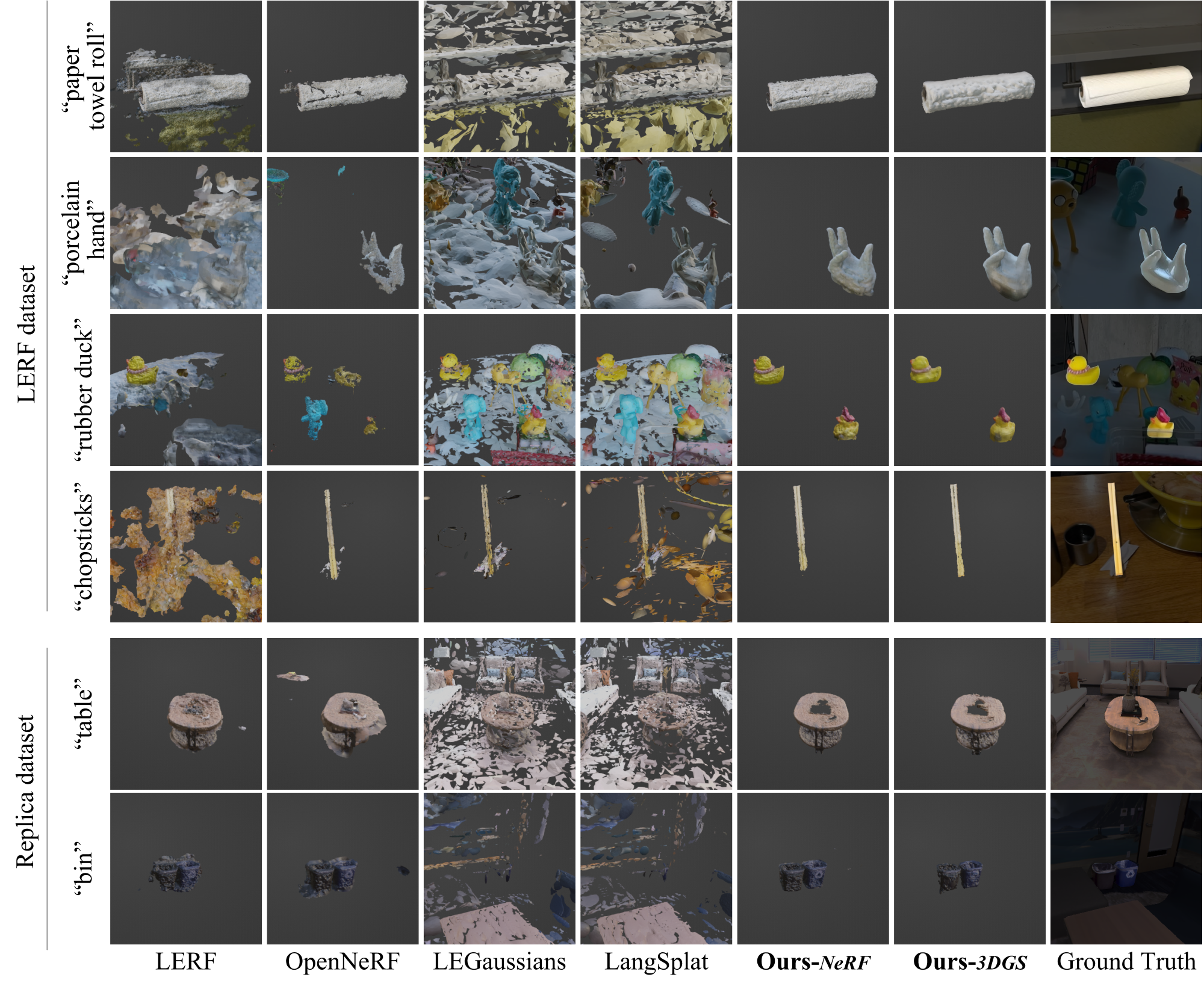}
  \caption{Qualitative comparisons of 3D segmentation on LERF and Replica datasets:
  We show an exported mesh of 3D querying results for the given text query. Unlike competitors, our method produces more clear boundaries in 3D segmentation results. 
    }
  \label{fig:3dseg}
\end{figure*}


\begin{figure*}[tb]
  \centering
  \includegraphics[width=0.68\linewidth]{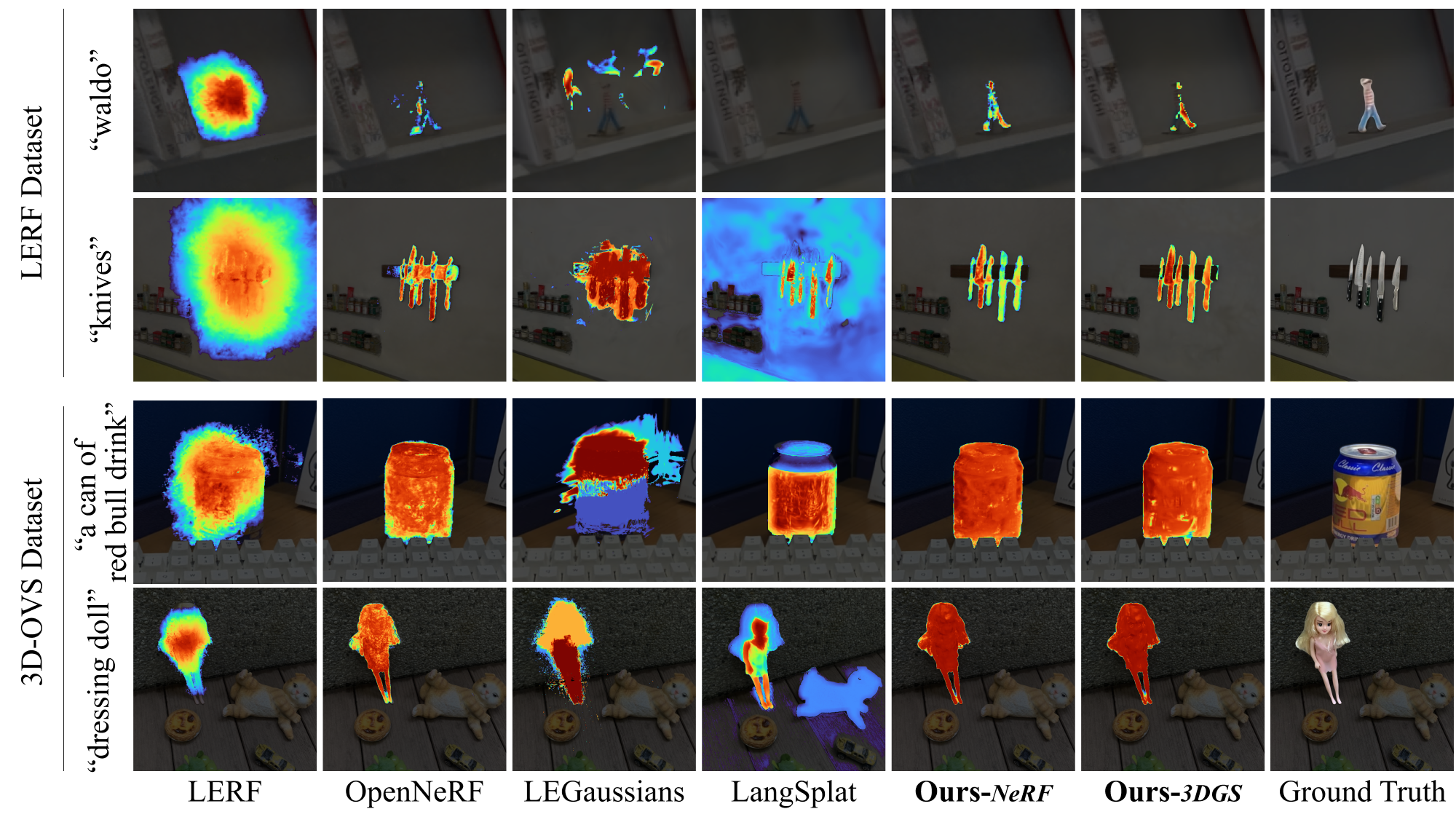}
  \caption{Comparisons of \PREVSEG on LERF and 3D-OVS datasets:
  We show a heatmap of the similarity for the given text query. We dim the background except for the target object, for better visualization. Our method achieves accurate segmentation results compared to competitors.
  }
  \label{fig:2dseg}
\end{figure*}


\subsection{3D Semantic Evaluation Protocol}
\label{sec:3dseg}
A previous work~\cite{engelmann2024opennerf} measures the mIoU between the prediction and annotation on the ground truth point cloud.
It might incorrectly classify the points with zero density and a correct embedding,
shown in \fref{fig:3d-seg-method}-(a).

To address this problem, we propose to measure the F1-score between
the exported mesh from the segmented results to ground truth mesh, inspired by surface reconstruction literature \cite{Knapitsch2017,li2023neuralangelo}.
Precision computes the ratio of correct volume among the estimated volume. Recall computes the ratio of covered volume among the GT volume. As it is impractical to compute the volume, we approximate the volume by a regular grid points in the estimated and GT meshes.
A point is considered \textit{correct} or \textit{covered} if there exists a point within a radius in the counterpart.
We note that this protocol can be generally applied to various neural representations, such as NeRF or 3DGS.


\section{Experiments}
In this section, we evaluate our methods on various datasets and compare them to competitors in terms of 3D and \prevseg with given text queries. 
We choose LERF, OpenNeRF, LEGaussians, and LangSplat as competitors, which are open-sourced. We re-evalute the competitors using the official code\footnote{The official implementation of LangSplat includes evaluation views during training, while we exclude them for fairness, following LERF and OpenNeRF.}.
Appendix provides details of the datasets.


\begin{table}[tb]
        \centering
        \begin{tabular}{c|l|c}
        \toprule
        \multirow{2}{*}{Backbone} & \multirow{2}{*}{Method} & Replica       \\
                                  &                         & F1-score~$\uparrow$     \\
                                  \midrule
        \multirow{3}{*}{NeRF}     & LERF                    & 0.0845                      \\
                                  & OpenNeRF                & 0.0361                     \\
                                  & \textbf{\ours-\textit{NeRF}}  & \textbf{0.1520}            \\
        \midrule
        \multirow{3}{*}{3DGS}     & LEGaussians              & 0.0067                     \\
                                  & LangSplat                & 0.0087                      \\
                                  & \textbf{\ours-\textit{3DGS}}                & \underline{0.1353}            \\
        \bottomrule
        \end{tabular}
        \caption{3D Segmentation Accuracy Comparison on the Replica dataset. \textbf{Bold} indicates the 1st, and \underline{underline} indicates the 2nd-best model.}
        \label{tab:segm:3D}
\end{table}

\paragraph{Evaluation} 
\label{sec:eval}
For quantitative comparison in 3D segmentation, we export meshes using the Poisson surface reconstruction. In LERF, OpenNeRF, and \ours-\textit{NeRF}, we use Nerfstudio~\cite{nerfstudio} to export meshes with 30K sampled points. In LEGaussians, LangSplat and \ours-\textit{3DGS}, we use SuGaR~\cite{guedon2023sugar} to export meshes. For \prevseg, we evaluate the mIoU and mAP between the ground truth mask and the predicted mask. For qualitative comparison in 3D segmentation, we use 50K points for Nerfstudio in LERF, OpenNeRF, and \ours-\textit{NeRF}.

\subsection{Segmentation Accuracy} 
\paragraph{3D Segmentation}
We compare qualitative results of 3D segmentation by exporting meshes for the region obtained through 3D querying (\sref{sec:querying}). In \tref{tab:segm:3D}, both of our models surpass the competitors. 

In \fref{fig:3dseg}, unlike competitors, \ours-\textit{NeRF} and \ours-\textit{3DGS} produce clear segmentation boundaries of the target object from the 3D scene. Notably, both of our models accurately segment complex shape objects and multiple objects like \texttt{porcelain hand} and \texttt{rubber ducks}. In the \texttt{rubber duck} query, LERF, LEGaussians, and LangSplat completely fail to localize the target object, while OpenNeRF incorrectly segments unrelated objects (e.g., a toy elephant and Jake) along with the target object. 


\begin{figure}[tb!]
  \centering
  \includegraphics[width=0.86\linewidth]{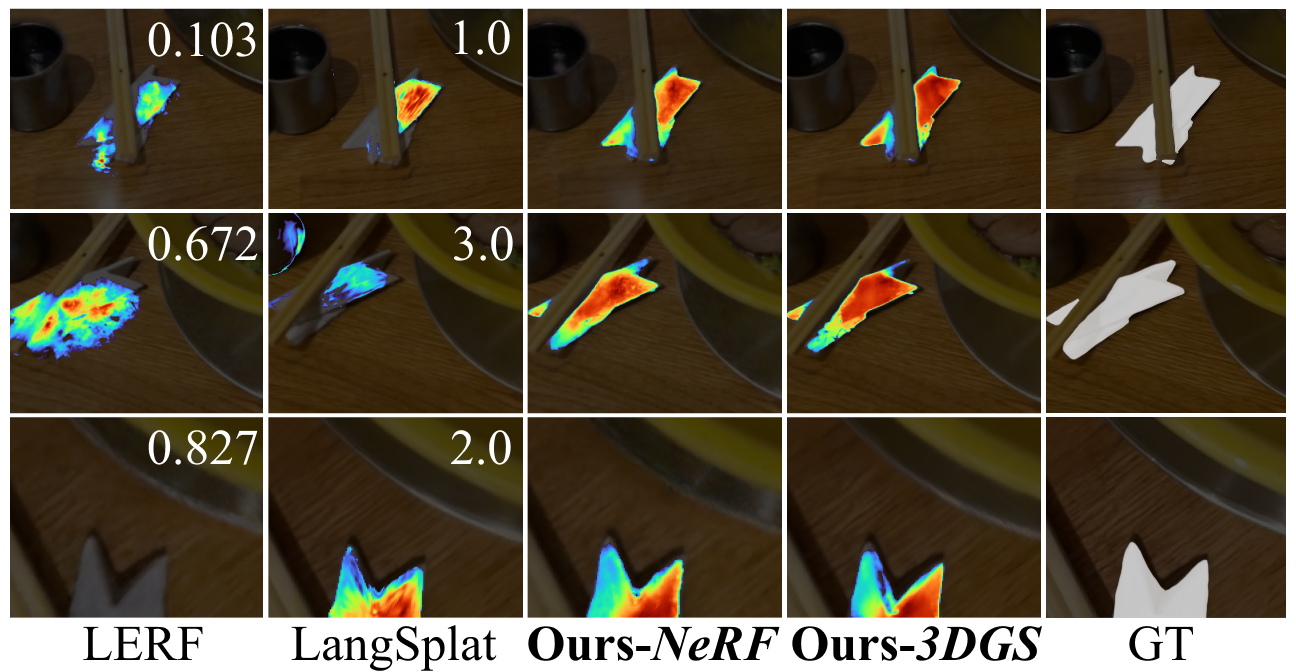}
  \caption{Comparison of view consistency: The figure shows the optimal scales of LERF and LangSplat for each viewpoint on \texttt{napkin} in LERF dataset, highlighting significant variation and view inconsistency in competitors. Our method avoids this by utilizing scale-free embeddings.
  }
  \label{fig:scale}
\end{figure}

\begin{table}[tb!] 
        \centering
        \begin{tabular}{l|cc|cc}
        \toprule
        \multirow{2}{*}{Method} & \multicolumn{2}{c|}{LERF}         & \multicolumn{2}{c}{3D-OVS}     \\
                                  & mIoU~$\uparrow$ & mAP~$\uparrow$    & mIoU~$\uparrow$ & mAP~$\uparrow$ \\
                                  \midrule
        LERF                    & 31.88          & 30.44            & 52.60          & 57.03         \\
        OpenNeRF                & 26.52          & 25.84            & 75.12          & 75.35         \\
        \textbf{\ours-\textit{NeRF}}           & \textbf{46.37}    & \textbf{45.86}      & \underline{77.46}   & \underline{84.65}  \\
        \midrule
        LEGaussians              & 21.43          & 20.81            & 47.98          & 50.86         \\
        LangSplat\footnotemark[3]               & 37.53          & 36.39            & 74.54          & 79.49         \\
        \textbf{\ours-\textit{3DGS}}      & \underline{44.37}   & \underline{44.57}     & \textbf{77.51}    & \textbf{84.86}   \\
        \bottomrule
        \end{tabular}
        \caption{Quantitative Results of \PREVSEG on LERF and 3D-OVS datasets.}
        \label{tab:segm:2D}
\end{table}


\paragraph{\PREVSEG} In \tref{tab:segm:2D}, both of our models show the highest \prevseg performance on LERF and 3D-OVS datasets. Also, we present qualitative results of \prevseg by comparing a heatmap for given text queries through 2D querying (\sref{sec:querying}). As shown in \fref{fig:2dseg}, \ours-\textit{NeRF} and \ours-\textit{3DGS} show clear segmentation boundaries. LangSplat and LEGaussians occasionally fail to find the target object, as seen with \texttt{waldo}. 

\fref{fig:scale} shows relevancy maps and optimal scales along different viewpoints. The float value on the top right of the image represents the optimal scale for LERF and LangSplat from each viewpoint. In \fref{fig:scale}, our method renders view consistent relevancy maps with \texttt{napkin} query. However, LERF and LangSplat show view inconsistent segmentation results due to the changing optimal scale. 


\begin{table*}[thb!]
\centering
\begin{tabular}{c|l|c|ccc|cc}
\toprule
\multirow{2}{*}{Backbone} & \multirow{2}{*}{Method} & \multirow{2}{*}{Training~$\downarrow$} & \multicolumn{5}{c}{Rendering}           \\ \hhline{~|~|~|-----}
                        &&                                & Render~$\downarrow$      & Decode~$\downarrow$       & Post-Process~$\downarrow$ & FPS~$\uparrow$  & Real-Time                           \\
                        \midrule
\multirow{3}{*}{NeRF} & LERF                    & 40 mins                        & 23688. ms    & -            & -               & 0.04 &\xmark  \\
& OpenNeRF                & 40 mins                       & 5944.7 ms    & -            & -               & 0.17 &\xmark \\
& \textbf{\ours-\textit{NeRF}}           & 40 mins                       & 2337.7 ms     & -            & -               & 0.43 &\xmark  \\
\midrule
\multirow{3}{*}{3DGS} & LEGaussians             & 90 mins                       & 15.665 ms    & 384.6 ms    & -               & \underline{2.50} &\xmark \\
& LangSplat               & 100 mins                      & 17.249 ms    & 0.935 ms     & 10473 ms        & 0.10 &\xmark \\ 
& \textbf{\ours-\textit{3DGS}}      & 40+5 mins                       & 14.257 ms    & -            & -               & \textbf{70.1} &\cmark\\
\bottomrule
\end{tabular}
\caption{Computational Time: We measure the computational cost on waldo\_kitchen scene on an RTX A5000.
}
\label{tab:cost}
\end{table*}


\begin{table*}[thb!]
    \centering
        \begin{tabular}{l|c|cc||ccc}
            \toprule
            \multirow{2}{*}{Method}       & Replica  & \multicolumn{2}{c||}{LERF} & \multicolumn{3}{c}{Computational Cost} \\
                                          & F1-score~$\uparrow$  & mIoU~$\uparrow$ & mAP~$\uparrow$& Training Time~$\downarrow$ & FPS~$\uparrow$  & \# of Gaussians~$\downarrow$  \\
            \midrule
            \textbf{\ours-\textit{3DGS}} & \textbf{0.1354} & \textbf{44.37} & \textbf{44.57} & \textbf{5 mins} & \textbf{70.1}  & \textbf{1M}\\
            w/o NeRF init                 & 0.1108          & 36.14          & 36.74 & 13 mins  & 41.2  & 2M        \\     
            \bottomrule
        \end{tabular}
    \caption{Ablation Study of Initializing NeRF into 3DGS:
    The ablation is conducted on Replica dataset for 3D segmentation and on LERF dataset for \prevseg.  We measure the computational cost on \textsl{waldo\_kitchen} scene with RTX A5000.}
    \label{tab:nerf_init}
\end{table*}


\subsection{Computational Time} 
\tref{tab:cost} shows the computational times of the LERF \textsl{waldo\_kitchen} scene with a single RTX A5000. We report the training time and rendering time for both ours and the competitors. 
LangSplat and LEGaussians train an autoencoder for each scene and require decoding during rendering. Note that LangSplat trains 3DGS from scratch, optimizes the language-embedded 3DGS per scale, heavily depends on post-processing (see Appendix).  
\ours-\textit{3DGS} takes 40 minutes to train \ours-\textit{NeRF}, 5 minutes to optimize 3DGS, and 22 milliseconds to query and transfer language field into 3DGS. \ours-\textit{3DGS} achieves real-time rendering of language embeddings for the first time, 28$\times$ faster than LEGaussians. 

\begin{table}[tb] 
    \centering
        \begin{tabular}{l|c|cc}
            \toprule
            \multirow{2}{*}{Method}       & Replica  & \multicolumn{2}{c}{LERF} \\
                                          & F1-score~$\uparrow$  & mIoU~$\uparrow$ & mAP~$\uparrow$  \\
            \midrule
            \textbf{Ours full} & \textbf{0.1532} & \textbf{46.37} & \textbf{45.86} \\
            w/o $L_\text{PS}$          & 0.0537          & 44.50          & 44.05          \\ 
            w/o $F_{\text{lang}}(\pos_i)$              & 0.1114          & 30.03          & 28.85          \\ 
            \bottomrule
        \end{tabular}
    \caption{Ablation Study of Supervising Semantics in 3D Space:
    The datasets are Replica and LERF for 3D and \prevseg, respectively. w/o $L_\text{PS}$ denotes using LERF CLIP loss $L_\text{lang}$ instead of \loss and w/o $F_{\text{lang}}(\pos_i)$ denotes using multi-scale embedding $F_{\text{lang}}(\pos_i, s)$ instead of scale-free embedding.
    }
    \label{tab:abl}
\end{table}

\subsection{Ablation Study}

\begin{figure}[tb]
  \centering
  \includegraphics[width=0.95\linewidth]{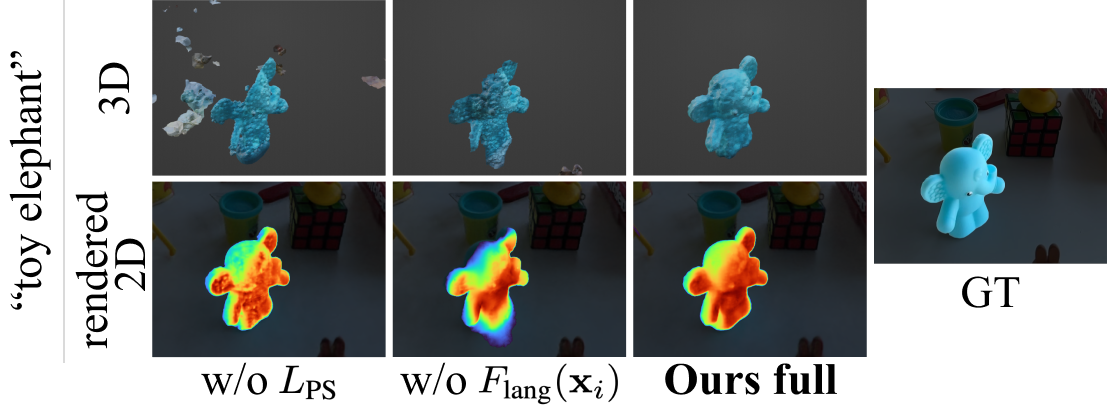}
  \caption{Qualitative Results of Ablation Study on \textsl{figurines} scene in LERF dataset.
  }
  \label{fig:ablation}
\end{figure}

\paragraph{Point-wise Semantic Loss}
We demonstrate the necessity of our \loss.
\fref{fig:ablation} qualitatively shows improvements due to \loss compared to 2D supervision with \eref{eq:2dloss}. Point-wise semantic loss removes meaningless floaters related to depth ambiguity.
\tref{tab:abl} shows that leveraging \loss instead of \eref{eq:2dloss} makes better understanding in 3D space, improving the F1-score (+0.0995). Furthermore, \loss also indirectly improves \prevseg accuracy. 

\paragraph{Scale-Free Embedding}
Our scale-free language embedding field from SAM-segmented objects produce greatly improved accuracy with clear object boundaries compared to multi-scale embedding field. \fref{fig:ablation} shows that scale-free embedding helps covering the head of the toy elephant which was lost with multi-scale embedding. In \tref{tab:abl}, the quantitative performance of the 3D and \prevseg greatly improves using scale-free embedding.

\paragraph{NeRF Initialization}
\label{sup:nerf_init}
We demonstrate the necessity of initializing 3DGS with the extracted point clouds from our learned NeRF. As shown in \tref{tab:nerf_init}, training 3DGS with SfM \cite{schoenberger2016sfm} initialization with the densification and pruning leads to performance degradation in both 3D and \prevseg. This indicates that aligning geometry and semantics is essential for transferring the language field to 3DGS. Moreover, NeRF initialization produces fewer Gaussians (0.50$\times$) which lead to less training time (0.38$\times$) and faster rendering speed (1.70$\times$).

\section{Conclusion}
We revisit the current literature on 3D understanding of NeRF and 3DGS and revise the problem setting. We reformulate the task to produce 3D segmented volumes instead of rendered 2D masks and propose a 3D evaluation protocol. We achieve state-of-the-art segmentation accuracy in both 3D and rendered 2D by computing loss directly on 3D points. 
Moreover, we enable the first real-time rendering speed among open-vocabulary methods by transferring the learned language field to 3DGS. 
We hope this paper drives forward a better 3D understanding of radiance fields by reconsidering the problem set.

\section*{Acknowledgements}
This work was supported by Institute of Information \& communications Technology Planning \& Evaluation(IITP) grant funded by the Korea government(MSIT) (No. RS-2024-00439762, Developing Techniques for Analyzing and Assessing Vulnerabilities, and Tools for Confidentiality Evaluation in Generative AI Models)

\bibliography{aaai25}

\iftrue
\appendix
\clearpage
\setcounter{figure}{4}
\setcounter{table}{8}
\setcounter{equation}{5}

\begin{table}[htb]
    \centering
    \begin{tabular}{l|ccc}
    \toprule
     Method     & Replica & LERF & 3D-OVS  \\
     \midrule
     LERF       & 0.50     & 0.50  & 0.50     \\
     OpenNeRF   & 0.50     & 0.50  & 0.50     \\
     \ours-\textit{NeRF}   & 0.55     & 0.60  & 0.55     \\
     \midrule
     LEGaussians   & 0.50     & 0.50  & 0.50     \\
     LangSplat   & 0.40     & 0.40  & 0.40     \\
     \ours-\textit{3DGS}   & 0.60     & 0.60  & 0.55     \\
     \bottomrule
    \end{tabular}
    \caption{Selected Thresholds in Quantitative Results.}
    \label{tab:threshold}
\end{table}


\begin{table*}
\begin{tabular}{cc} 
\toprule
\multirow{3}{*}{ \quad \textsl{figurines} \qquad }        & green apple, ice cream cone, jake, miffy, old camera, pikachu, pink ice cream, \\
                                  & porcelain hand, quilted pumpkin, rabbit, red apple, rubber duck, rubics cube, spatula, \\
                                  & tesla door handle, toy cat statue, toy chair, toy elephant, twizzlers, waldo \\ 
\midrule
\multirow{2}{*}{ \quad \textsl{ramen} \qquad }            & bowl, broth, chopsticks, egg, glass of water, green onion, napkin, nori, \\
                                  & pork belly, ramen, sake cup, wavy noodles \\ 
\midrule
\multirow{3}{*}{ \quad \textsl{teatime} \qquad }          & bag of cookies, bear nose, coffee, coffee mug, cookies on a plate, \\
                                  & dall-e, hooves, paper napkin, plate, sheep, spill, spoon handle, \\
                                  & stuffed bear, tea in a glass, yellow pouf \\ 
\midrule
\multirow{3}{*}{ \quad \textsl{waldo\_kitchen} \qquad }    & blue hydroflask, coffee grinder, cookbooks, cooking tongs, copper-bottom pot, \\
                                  & dish soap, faucet, knives, olive oil, paper towel roll, pepper mill, pour-over vessel, \\
                                  & power outlet, red mug, scrub brush, sink, spice rack, utensils, vegetable oil, waldo \\
\bottomrule
\\
\end{tabular}
\caption{Text Query of LERF Dataset.}
\label{tab:text_query}
\end{table*} 

\section{Experiments Details}
\subsection{Implementation Details}
We train the NeRF and the CLIP branch jointly from scratch with the same configurations of LERF except the learning rate on Replica datasets which is 1e-4. The CLIP features are extracted using OpenCLIP ViT-B/16 trained on LAION-2B. We use an automatic mask generator from ViT-H SAM with the default configuration. 

We optimize 3DGS in \sref{sec:3dgs_lang} with the same settings as the official 3DGS, except for the following modifications: 1) freezing the position of the Gaussian center $\pos$, 2) setting the rotation learning rate to 1e-4, and 3) running for 10K iterations without a densification process. 

We train our models and competitors with a single RTX A5000.

\begin{figure}[bh!]
  \centering
  \includegraphics[width=0.8\linewidth]{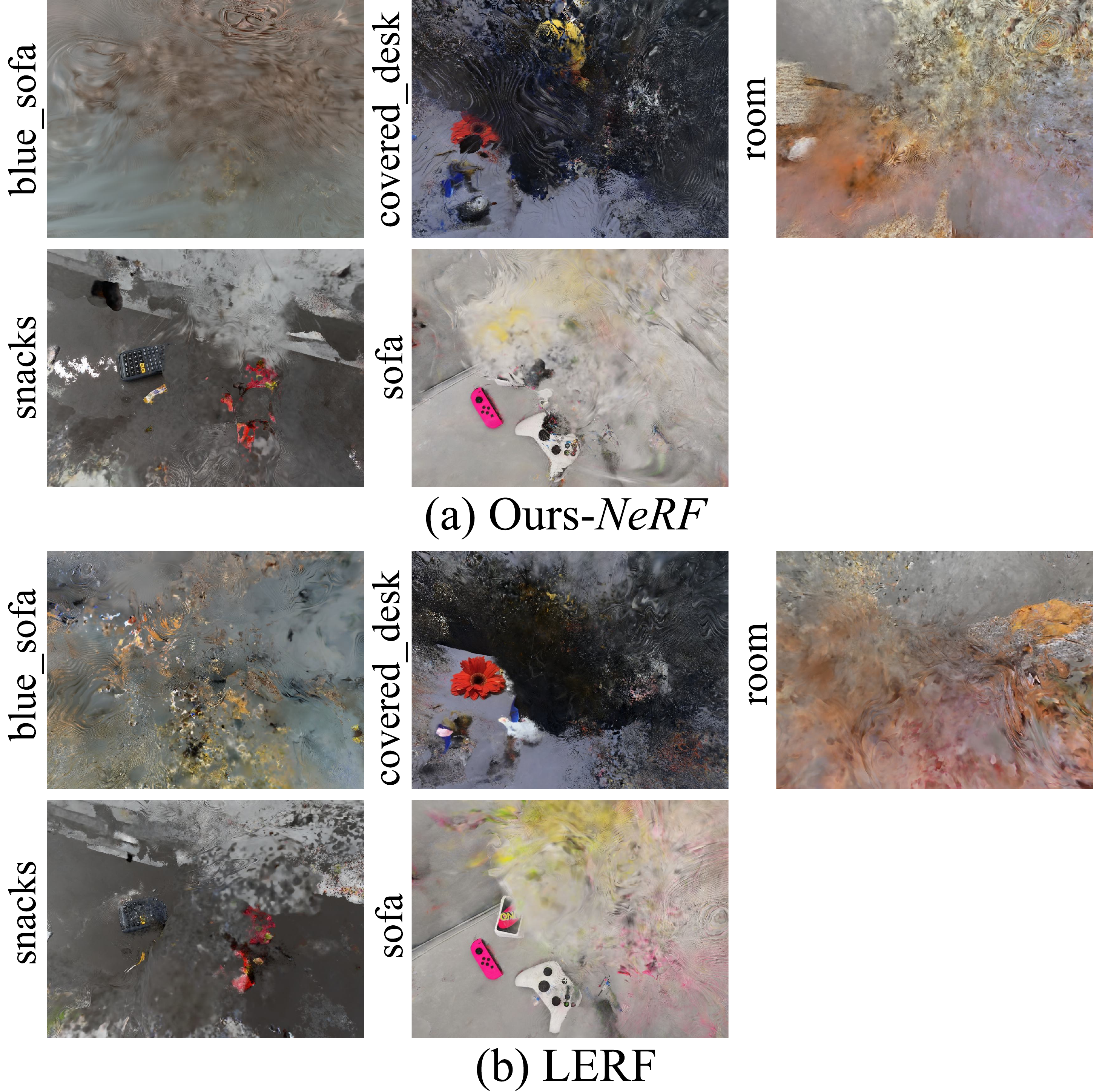}
  \caption{Poor reconstruction quality in evaluation view of 3D-OVS dataset.}
  \label{fig:3dovs-recon}
\end{figure}
\subsection{Details of Datasets}
\label{sup:dataset}
We use three datasets for evaluation: Replica dataset for 3D segmentation, and LERF and 3D-OVS datasets for \prevseg. All of these datasets are widely used to evaluate the segmentation performance of radiance fields. We fix the thresholds for each dataset as detailed in \tref{tab:threshold} to evaluate the performance.
\paragraph{Replica dataset.} Replica dataset is a high-quality indoor 3D mesh dataset with semantic labels. We measure 3D segmentation performance with labeled 3D mesh, using class names as the text queries. 

\paragraph{LERF dataset.} LERF dataset is a collection of real-world scenes captured by iPhone. This includes text queries and their ground truth 2D bounding boxes. In LERF dataset, most competitors~\cite{qin2023langsplat,shi2023language} report their performance by building their own ground truth mask datasets corresponding to their partly chosen text queries. However, it is a challenge to compare because the performance of open-vocabulary segmentation can significantly vary depending on the text query. 

Hence, starting from the given text queries and 2D bounding boxes, we use SAM to create auxiliary masks and manually correct errors. SAM\cite{kirillov2023segany} receives an image and a user prompt (bounding box or point) as input and creates three masks. Among these candidate masks, we select the ground truth mask that corresponds most closely to the queried object.
Then, we manually label a few inaccurate masks. The masks will be publicly available.
We use four scenes \textsl{figurines}, \textsl{ramen}, \textsl{teatime}, \textsl{waldo\_kitchen} from LERF dataset.
We choose three evaluation views that are not blurry. These masks will be publicly available online. 

\paragraph{3D-OVS dataset.} 3D-OVS dataset is a collection of real-world scenes, text queries, and their ground truth pixel-level labels. The dataset consists of 10 scenes, each with 20-30 viewpoints, including five evaluation viewpoints. Several scenes in 3D-OVS dataset show poor reconstruction quality from the evaluation view shown in \fref{fig:3dovs-recon}. Therefore, we report the performance only for the scenes with successful reconstruction: (\textsl{bed}, \textsl{bench}, \textsl{lawn}, and \textsl{office\_desk}).

\paragraph{ScanNet dataset.} ScanNet dataset is a real-world dataset that includes RGB-D scans, meshes, and instance-level semantic labels. It is widely used to evaluate performance in 3D scene understanding tasks. In our case, we select one scene (\textsl{scene0003}) and provide the 3D segmentation accuracy, in \aref{sup:more}.

\section{Details of Previous Language Embed 3DGS}

\subsection{Splatting High Dimensions Embeddings on 3DGS}
\label{sup:gpu}
To jointly train 3DGS with language features, a straightforward approach is to embed the 512-dimensional language feature into each Gaussian. However, this approach results in out-of-memory(OOM) issues. 
Splatting 512-dimensional language embeddings requires 522 KB of shared memory per streaming multiprocessor (SM). Specifically, 3DGS \cite{kerbl3Dgaussians} requires 10 KB of shared memory and needs an additional 1 KB/SM for each increase in the dimension of the language feature. 3DGS allocates static shared memory, and CUDA limits this allocation to 48 KB for architectural compatibility. Even with a dynamic allocation that enables allocation above the 48 KB limit, 522 KB of shared memory exceeds the maximum shared memory of current top GPUs (e.g., the H100 has a maximum shared memory of 228 KB).

Consequently, existing methods address these challenges using the following strategies, each with its trade-offs:
\begin{itemize}
    \item \textbf{Compress high-dimensional features}~\cite{qin2023langsplat,shi2023language}. This approach accelerates training but reduces segmentation performance and requires an additional decoding process.
    \item \textbf{Allocate rasterizer variables for the backward pass to global memory instead of shared memory}~\cite{zhou2023feature}.  While this avoids performance degradation, it significantly slows down training.
    \item \textbf{Use dynamic shared memory allocation with reduced tile size in the rasterizer}~\cite{ye2024gsplat}. This method is faster than global memory allocation while using more shared memory, but it still slows down training due to the smaller tile size.
    \item \textbf{Render in chunks}~\cite{bhalgat2024n2f2}. Although this accelerates training and prevents performance degradation, it drops fps (10 fps reported in the original paper).
\end{itemize}
Compressing the features produces poor mIoU (-7.30) but speeds up the training time 47$\times$, as shown in \tref{tab:tradeoff}. However, \ours-\textit{3DGS} achieves fast training without feature compression.

\begin{table}[hbt!]
    \centering
    \begin{tabular}{l|cc|c}
        \toprule
        Method         & mIoU~$\uparrow$ & mAP~$\uparrow$  & Training Time~$\downarrow$      \\
        \midrule
        3dim-feature          & 31.54 & 30.09 & \textbf{30 mins}           \\
        512dim-feature        & \textbf{38.84} & \textbf{38.20} & 1400 mins          \\
        \bottomrule
        \end{tabular}
    \caption{Trade-off of Feature Compression on LERF dataset: we compress the feature following LangSplat while training each model with the same number of Gaussians for a fair comparison.}
    \label{tab:tradeoff}
\end{table}

\subsection{Post-process of LangSplat}
\label{sup:postprocess}
We show the performance difference with and without post-processing in LangSplat, as shown in \fref{fig:post-process}.
Without post-processing, it achieves a fast rendering speed of 55 fps (550$\times$)~\footnote{The post-processing consists of a mean convolution filter on the relevancy map implemented with OpenCV (61.98ms) and a mean convolution filter on the binary mask implemented with nested for-loops (3249 ms). This is done for three scales, 10473 ms in total.} but produces a poor mIoU (-5.5) as shown in \tref{tab:langsplat_postprocess}. Note that \ours-\textit{3DGS} achieve faster rendering speed even compared to LangSplat without post-processing, as shown in \tref{tab:cost}.

\begin{table}[hbt]
\centering
\begin{tabular}{l|cc}
\toprule
LangSplat           & mIoU$\uparrow$  & mAP$\uparrow$   \\
\midrule
w/ post-process  & \textbf{37.53} & \textbf{36.39} \\
w/o post-process & 32.01 & 30.64 \\
\bottomrule
\end{tabular}
\caption{Post-process of LangSplat on LERF dataset.}
\label{tab:langsplat_postprocess}
\end{table}

\begin{figure}[hbt]
  \centering
  \includegraphics[width=0.8\linewidth]{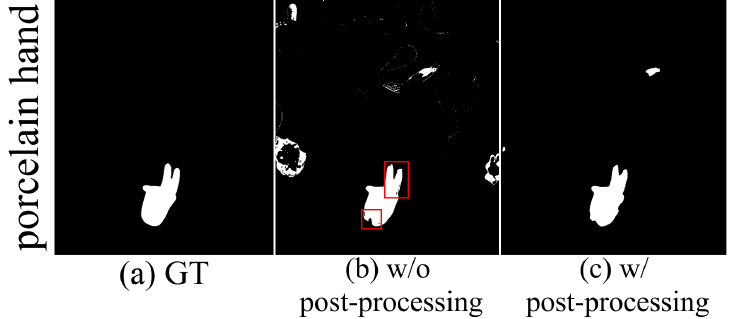}
  \caption{LangSplat includes post-processing, which significantly improves performance by reducing floaters.
  }
  \label{fig:post-process}
\end{figure}

\section{More Results}
\label{sup:more}
We mainly evaluate 3D segmentation accuracy on Replica datasets. To expand this, we also evaluate and compare on \texttt{scene\_0003} of ScanNet~\cite{dai2017scannet} dataset, in \tref{tab:scannet}.

Additionally, we report the quantitative results of each scene on Replica dataset, LERF dataset, and 3D-OVS dataset, in \tref{tab:per_replica}-\ref{tab:per_3dovs}.

\begin{table*}[tb]
        \centering
        \begin{tabular}{c|l|c}
        \toprule
        \multirow{2}{*}{Backbone} & \multirow{2}{*}{Method} & ScanNet       \\
                                  &                         & F1-score~$\uparrow$     \\
                                  \midrule
        \multirow{3}{*}{NeRF}     & LERF                    & 0.0333                      \\
                                  & OpenNeRF                & 0.0663                     \\
                                  & \textbf{\ours-\textit{NeRF}}  & \textbf{0.0849}            \\
        \midrule
        \multirow{3}{*}{3DGS}     & LEGaussians              & 0.0064                     \\
                                  & LangSplat                & 0.0340                      \\
                                  & \textbf{\ours-\textit{3DGS}}                & \underline{0.0769}            \\
        \bottomrule
        \end{tabular}
        \caption{3D Segmentation Accuracy Comparison on ScanNet dataset.}
        \label{tab:scannet}
\end{table*}

\begin{table*}[htb!]
\centering
\begin{tabular}{c|l|c|c|c|c}
\toprule 
\multirow{2}{*}{Backbone} & \multirow{2}{*}{Method} & \textsl{office\_0} & \textsl{office\_1} & \textsl{office\_2} & \textsl{office\_3} \\
                        && F1-score$\uparrow$ & F1-score$\uparrow$ & F1-score$\uparrow$ & F1-score$\uparrow$ \\
                        \midrule
\multirow{3}{*}{NeRF} & LERF                    & 0.0774      & 0.0191             & 0.0644             & 0.0621         \\
&OpenNeRF                & 0.0486             & 0.0169             & 0.0318             & 0.0356         \\
&\ours-\textit{NeRF}    & \textbf{0.1434}       & \textbf{0.0434}       & \textbf{0.1493}       & \textbf{0.1255}   \\
\midrule
\multirow{3}{*}{3DGS} &LEGaussians             & 0.0013             & 0.0043             & 0.0109             & 0.0031         \\
&LangSplat               & 0.0004             & 0.0123             & 0.0044             & 0.0038         \\
&\textbf{\ours-\textit{3DGS}}    & \underline{0.1097}             & \underline{0.0406}      & \underline{0.1395}      & \underline{0.1181}  \\
\midrule
\midrule
\multirow{2}{*}{Backbone} & \multirow{2}{*}{Method} & \textsl{office\_4} & \textsl{room\_0} & \textsl{room\_1} & \textsl{room\_2} \\
                        && F1-score$\uparrow$ & F1-score$\uparrow$ & F1-score$\uparrow$ & F1-score$\uparrow$ \\
                        \midrule
\multirow{3}{*}{NeRF} &LERF                    & 0.0645             & 0.1124             & \underline{0.1607}      & 0.1158         \\
&OpenNeRF                & 0.0521             & 0.0180             & 0.0432             & 0.0423         \\
&\ours-\textit{NeRF}    & \textbf{0.1722}       & \textbf{0.1654}       & \textbf{0.1934}       & \textbf{0.2232}   \\
\midrule
\multirow{3}{*}{3DGS} &LEGaussians             & 0.0111             & 0.0080             & 0.0066             & 0.0085         \\
&LangSplat               & 0.0107             & 0.0139             & 0.0122             & 0.0118         \\
&\textbf{\ours-\textit{3DGS}}    & \underline{0.1629}     & \underline{0.1636}      & 0.1529             & \underline{0.1959}  \\
\bottomrule
\end{tabular}
\caption{Per-scene quantitative results on Replica dataset.}
\label{tab:per_replica}
\end{table*}

\begin{table*}[htb!]
\centering
\begin{tabular}{l|cc|cc|cc|cc}
\toprule
\multirow{2}{*}{Method} & \multicolumn{2}{c|}{\textsl{figurines}} & \multicolumn{2}{c|}{\textsl{ramen}} & \multicolumn{2}{c|}{\textsl{teatime}} & \multicolumn{2}{c}{\textsl{waldo\_kitchen}} \\
                        & mIoU$\uparrow$    & mAP$\uparrow$ & mIoU$\uparrow$    & mAP$\uparrow$ & mIoU$\uparrow$    & mAP$\uparrow$ & mIoU$\uparrow$    & mAP$\uparrow$ \\
                        \midrule
LERF                    & 40.67             & 38.75         & 24.56             & 24.35         & 36.85             & 35.06         & 25.44             & 23.62         \\
OpenNeRF                & 16.24             & 16.02         & 23.05             & 23.57         & 39.08             & 38.26         & 27.69             & 25.50         \\
\textbf{\ours-\textit{NeRF}}    & \textbf{59.79}       & \textbf{59.00}   & 37.44             & 38.38         & \textbf{47.59}       & \textbf{46.92}   & \underline{40.67}      & \underline{39.16}  \\
\midrule
LEGaussians             & 21.02             & 20.57         & 24.90             & 24.67         & 24.50             & 24.09         & 15.28             & 13.89         \\
LangSplat               & 42.23             & 40.55         & \textbf{48.17}       & \textbf{48.71}   & 37.74             & 36.26         & 21.98             & 20.05         \\
\textbf{\ours-\textit{3DGS}}    & \underline{56.23}      & \underline{55.79}  & \underline{38.28}      & \underline{40.81}  & \underline{41.59}      & \underline{41.16}  & \textbf{42.60}       & \textbf{41.79}   \\
\bottomrule
\end{tabular}
\caption{Per-scene quantitative results on LERF dataset.}
\label{tab:per_lerf}
\end{table*}

\begin{table*}[htb!]
\centering
\begin{tabular}{l|cc|cc|cc|cc}
\toprule
\multirow{2}{*}{Method} & \multicolumn{2}{c|}{\textsl{bed}} & \multicolumn{2}{c|}{\textsl{bench}} & \multicolumn{2}{c|}{\textsl{lawn}} & \multicolumn{2}{c}{\textsl{office\_desk}} \\
                        & mIoU$\uparrow$    & mAP$\uparrow$ & mIoU$\uparrow$    & mAP$\uparrow$ & mIoU$\uparrow$    & mAP$\uparrow$ & mIoU$\uparrow$    & mAP$\uparrow$ \\
                        \midrule
LERF                    & 45.29             & 57.74         & 50.88             & 51.59         & 54.19             & 57.40         & 60.04             & 61.41         \\
OpenNeRF                & \textbf{82.25}       & \textbf{82.40}   & \underline{70.24}      & 71.14         & 78.70             & 79.27         & 69.29             & 69.79         \\
\textbf{\ours-\textit{NeRF}}    & 66.71             & 78.65         & 66.69             & 78.81         & \textbf{96.74}       & \textbf{96.98}   & \textbf{79.72}       & \underline{84.18}  \\
\midrule
LEGaussians             & 33.23             & 44.94         & 63.86             & 64.25         & 43.48             & 44.20         & 51.36             & 50.05         \\
LangSplat               & 65.74             & 73.01         & \textbf{82.75}       & \textbf{88.06}   & 82.41             & 87.95         & 67.26             & 68.93         \\
\textbf{\ours-\textit{3DGS}}    & \underline{67.40}      & \underline{79.33}  & 67.08             & \underline{79.20}  & \underline{96.25}      & \underline{96.92}  & \underline{78.89}      & \textbf{84.36}   \\
\bottomrule
\end{tabular}
\caption{Per-scene quantitative results on 3D-OVS dataset.}
\label{tab:per_3dovs}
\end{table*}
\fi

\end{document}